\documentclass{bmvc2k}
\usepackage{stfloats}
\usepackage{color}
\usepackage{amssymb}
\usepackage{bbding}
\usepackage{multirow,booktabs}
\usepackage{bm}

\title{Disparity Estimation Using a Quad-Pixel Sensor}

\addauthor{Zhuofeng Wu}{zwu@ok.sc.e.titech.ac.jp}{1}
\addauthor{Doehyung Lee}{dlee@ok.sc.e.titech.ac.jp}{1}
\addauthor{Zihua Liu}{zliu@ok.sc.e.titech.ac.jp}{1}
\addauthor{Kazunori Yoshizaki}{kazunori.yoshizaki@olympus.com}{2}
\addauthor{Yusuke Monno}{ymonno@ok.sc.e.titech.ac.jp}{1}
\addauthor{Masatoshi Okutomi}{mxo@ctrl.titech.ac.jp}{1}

\addinstitution{
 Institute of Science Tokyo\\
 Tokyo, Japan
}
\addinstitution{
Olympus Medical Systems Corporation\\
Tokyo, Japan
}

\runninghead{Wu et al.}{Disparity Estimation Using a Quad-Pixel Sensor}

\begin{document}

\maketitle
\begin{abstract}

A quad-pixel~(QP) sensor is increasingly integrated into commercial mobile cameras. The QP sensor has a unit of 2$\times$2 four photodiodes under a single microlens, generating multi-directional phase shifting when out-focus blurs occur. Similar to a dual-pixel~(DP) sensor, the phase shifting can be regarded as stereo disparity and utilized for depth estimation. Based on this, we propose a QP disparity estimation network~(QPDNet), which exploits abundant QP information by fusing vertical and horizontal stereo-matching correlations for effective disparity estimation. We also present a synthetic pipeline to generate a training dataset from an existing RGB-Depth dataset.  Experimental results demonstrate that our QPDNet outperforms state-of-the-art stereo and DP methods. Our code and synthetic dataset are available at \href{https://github.com/Zhuofeng-Wu/QPDNet}{https://github.com/Zhuofeng-Wu/QPDNet}.
\end{abstract}


\section{Introduction}
\label{sec:intro}

\begin{figure*}
\begin{center}
\centerline{\includegraphics[width=13cm]{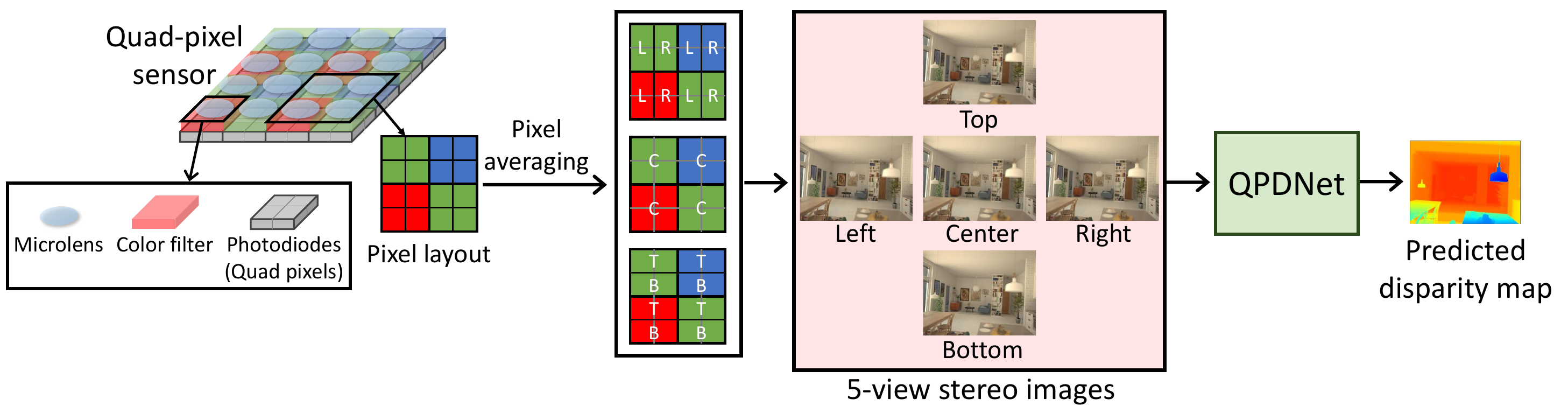}}
\end{center}
\vspace{-11mm}
   \caption{The overall flow of our disparity estimation. From the data captured by a QP sensor, we generate five-view stereo images, where disparities exist between a reference center image and each of the left, the right, the top, and the bottom images, according to the phase shifting principle of the DP/QP sensor in out-focus regions~\cite{Okawa2019quad}. Using those five-view images as inputs, our QPDNet predicts a disparity map aligned to the reference center image.}
\label{fig:overview}
\end{figure*}

Recent advancements in sensor technology have led to the gradual integration of dual-pixel~(DP)~\cite{garg2019learning} and quad-pixel~(QP)~\cite{Okawa2019quad} sensors into commercial digital cameras and smartphones. As shown in Fig.~\ref{fig:overview}, a QP sensor has a unit of 2$\times$2 photodiodes under a single microlens and an identical color filter. Akin to a DP sensor, where 1$\times$2 photodiodes are used, the QP sensor generates phase shifting when out-focus blurs occur. The phase-shifting information is validated to be valuable in DP sensors and various applications, such as defocus debluring~\cite{abuolaim2020defocus}, reflection removal~\cite{punnappurath2019reflection}, and raindrop removal~\cite{li2022rain}, have been proposed.

Another high-potential application is depth estimation because the out-focus phase shifting can be regarded as stereo disparity. While the DP sensor only utilizes horizontal disparity in depth estimation~\cite{wadhwa2018synthetic, punnappurath2020modeling, Kim2023Dualpixel, garg2019learning}, the QP sensor can exploit multi-directional disparities using the 2$\times$2 photodiodes structure. Consequently, the QP sensor provides a functionality similar to a multi-baseline stereo~\cite{Okutomi1993Stereo} and a camera array~\cite{Holloway2015Spatial} only using a single sensor. Despite this potential, as far as we know, no literature has been published yet for disparity estimation using a QP sensor.

In this paper, we propose a QP disparity estimation network~(QPDNet), which effectively utilizes multi-directional disparities observed using a QP sensor. As shown in Fig.~\ref{fig:overview}, we first apply three kinds of pixel averaging of raw pixel values and generate five-view (i.e., left, right, center, top, and bottom) stereo images, where the center image is used as a reference to integrate two horizontal~(center-right and center-left) and two vertical~(center-top and center-bottom) disparity information. Inspired by the success of RAFT-Stereo~\cite{lipson2021raft}, we propose to construct four-directional correlation pyramids and effectively fuse them by introducing a novel multi-direction lookup fusion~(MDLF) module, which is designed to extract local correlations in multiple directions and to enable adaptive fusion of extracted correlations by utilizing both channel-wise and pixel-wise attention mechanisms. As in RAFT-Stereo, the local correlations output by the MDLF module along with the current disparity and context features are iteratively fed to gated recurrent units~(GRUs) to predict a final disparity map.

While some studies constructed datasets for depth estimation with DP sensors~\cite{abuolaim2020defocus, punnappurath2020modeling, wadhwa2018synthetic, garg2019learning}, they cannot be directly used for QP sensors, because only left and right image pairs are provided. Thus, we rebuild a synthetic pipeline to generate a QP image along with a ground-truth disparity map using the pair of an all-in-focus image and a ground-truth depth map available in existing RGB-D datasets~\cite{roberts2021}, by following the DP blur kernel and image formation modeling of~\cite{abuolaim2021learning}. The generated datasets are used for the training and the evaluation of our QPDNet.  

In summary, our main contributions are listed as follows:
\begin{itemize}
    \vspace{-2mm}
    \item We proposed QPDNet, a novel network to effectively predict a disparity map from QP data by combining horizontal and vertical disparities observed by a QP sensor.
    \vspace{-2mm}
    \item We proposed MDLF module, a novel module to effectively extract and fuse the multi-directional stereo-matching correlations.
    \vspace{-2mm}
    \item We constructed a dataset for disparity estimation from QP data based on a forward image formation process for the QP sensor.
    \vspace{-2mm}
    \item We experimentally validated that QPDNet outperforms other state-of-the-art stereo and DP methods in both synthetic and real experiments.
\end{itemize}

\section{Related Work}
\label{related_works}
\subsection{Stereo Disparity Estimation}
Given a pair of rectified stereo images, stereo matching aims to compute the disparity for each pixel in the reference image. Recently, CNN-based stereo methods \cite{ganet,psmnet,GOAT,NiNet,xu2023iterative}
have achieved impressive performance on most of the standard benchmarks. Most of them construct a cost volume in the feature level to measure the matching similarity. The pioneering work DispNetC \cite{dispnetc} utilizes a correlation layer to calculate the inner product of left and right features at each disparity level for measuring the similarity. GCNet~\cite{gcnet} pioneered the use of concatenated left and right image features to construct a 4D volume, enriching content representation for similarity assessment. This approach, featuring a concatenation volume followed by stacked 3D convolutional networks for aggregation, has been widely adopted in many contemporary state-of-the-art methods, such as \cite{psmnet,ganet,shen2021cfnet}. Building on this, GwcNet~\cite{gwcnet} and ACVNet~\cite{ACVNet} integrate the strengths of both correlation and concatenation volumes using a group-wise correlation strategy. Recent advancements \cite{STTR,GOAT,UnifyStereo} have incorporated attention mechanisms within Transformers, employing cross-attention \cite{chen2021crossvit} to replace traditional cost volumes. Although deep-learning-based stereo matching has achieved significant progress, the inherent nature of stereo images, which provide information from only two limited views and one matching direction, often leads to unsatisfactory results, as evidenced by that a multi-baseline stereo~\cite{Okutomi1993Stereo} improves the robustness.

\subsection{DP Disparity Estimation}
As DP sensors have become adopted in consumer cameras and smartphones, an increasing number of studies have emerged aiming to estimate depth/disparity from a DP image. As an early study, Wadhwa et al. adapted classical stereo-matching techniques to infer a disparity map from two DP views~\cite{wadhwa2018synthetic}. Punnappurath et al. modeled the defocus-disparity characteristic of the DP sensor and introduced a model-based optimization method to estimate the defocus map, which can be interpreted as an inverse depth map~\cite{punnappurath2020modeling}. Xin et al. proposed another technique for estimating a defocus map by optimizing a multiplane image representation to best explain the observed DP image using the calibrated point spread function~(PSF) kernels~\cite{xin2021defocus}. In recent years, learning-based methods have demonstrated superior performance. Garg et al. devised a learning-based approach for DP depth estimation, employing an affine invariant objective function to estimate inverse depth~\cite{garg2019learning}. Pan et al. introduced an end-to-end network capable of jointly estimating depth and producing a deblurred image~\cite{pan2021dual}. Kang et al. proposed a network to estimate the depth and surface normal focusing on faces~\cite{kang2022facial}.
Kim et al. proposed a self-supervised learning method leveraging the symmetric property of DP PSF kernels~\cite{Kim2023Dualpixel}.

The closest work to ours is the one by Zhang et al.~\cite{zhang20202}, which applies two vertically aligned cameras, where one of the cameras is a DP camera. This configuration provides vertical disparities as a standard stereo and horizontal disparities as a DP stereo. While this configuration can leverage the different baseline lengths of the standard stereo and the DP stereo, it requires the alignment between the stereo and the DP setups due to the necessity of stereo rectification for two cameras. In contrast, our configuration using a single QP camera eliminates the need for the rectification and the alignment to handle the vertical and the horizontal disparities by using the center image as a reference image, thus making the overall system and method simpler. 

\section{QP Disparity Estimation Dataset Generation}
\label{sec:syn_data}

\begin{figure*}
\begin{center}
\centerline{\includegraphics[width=13cm]{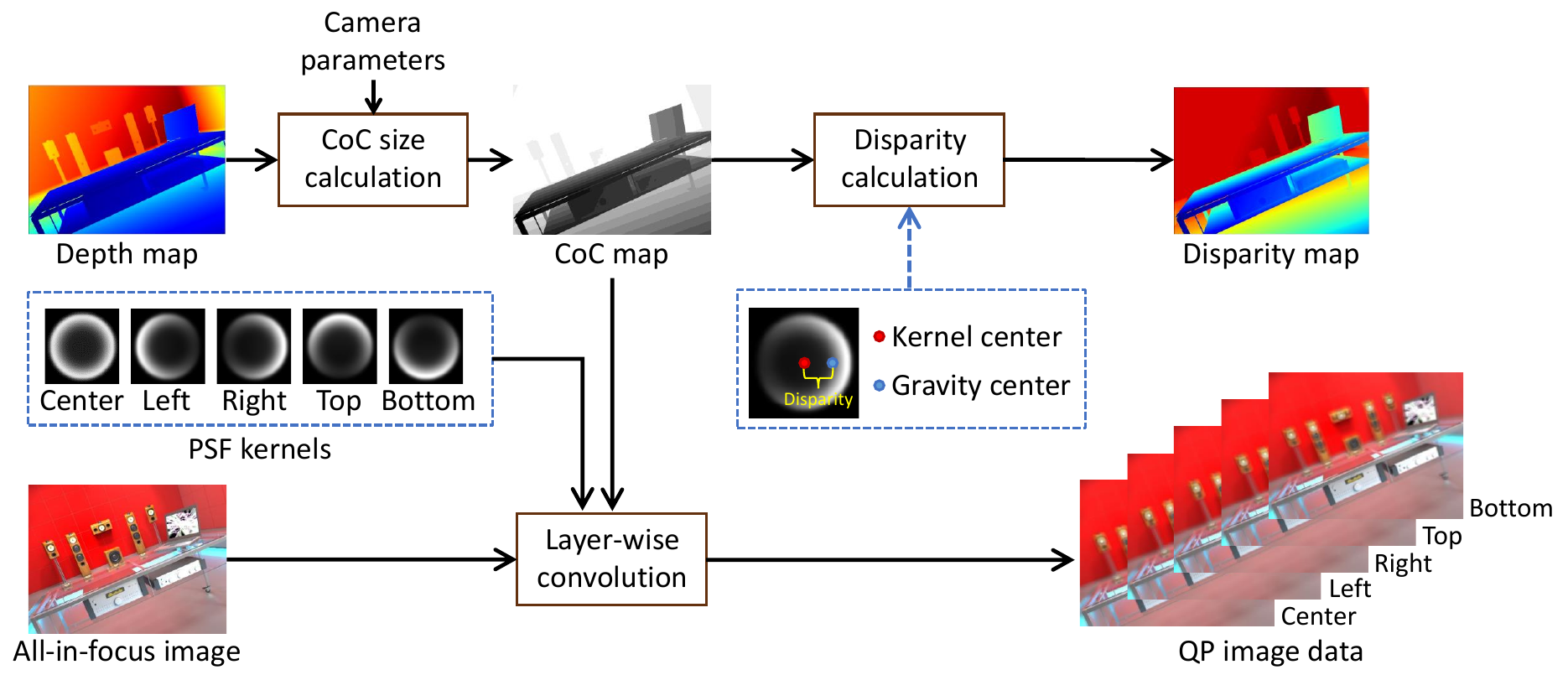}}
\end{center}
   \vspace{-10mm}
   \caption{The overview of our data generation process.}
\label{fig:syn_process}
\end{figure*}

Since a dataset specifically tailored for QP disparity estimation is currently unavailable, we design a synthetic data generation pipeline, as depicted in Fig.~\ref{fig:syn_process}. The inputs are the pair of a depth map and an all-in-focus image, which is widely available in existing RGB-D datasets~\cite{roberts2021, silberman2012indoor, zheng2020structured3d}. Using the depth map and the considered camera parameters, the circle-of-confusion~(CoC) radius of each pixel is calculated as
\begin{equation}
CoC = \frac{1}{p} \times \frac{f}{2F} \times \frac{f}{f_d - f} \times \frac{z - f_d}{z},
\label{formula: coc}
\end{equation}
where $z$ represents the depth. $f$, $f_d$, $F$, and $p$ denote the camera's focal length, focal distance, F-stop (aperture), and pixel size, respectively.

Subsequently, we adopt DP left and right PSF kernel models of~\cite{abuolaim2021learning} and rotate them by 90$^\circ$ to generate the PSFs for the top and the bottom images. The PSF for the center image is also generated by averaging the left and the right kernels. Then, the corresponding all-in-focus image is convolved with the PSFs of each image in a layer-wise manner of~\cite{lu2021self,lee2019deep} to generate a set of five images. Finally, we define the disparity as the distance between the kernel center and the gravity center of the right PSF and compute the disparity of each pixel with the corresponding CoC size.
\section{Proposed QPDNet}

\subsection{Network Architecture Overview}

\begin{figure*}
\begin{center}
\centerline{\includegraphics[width=13cm]{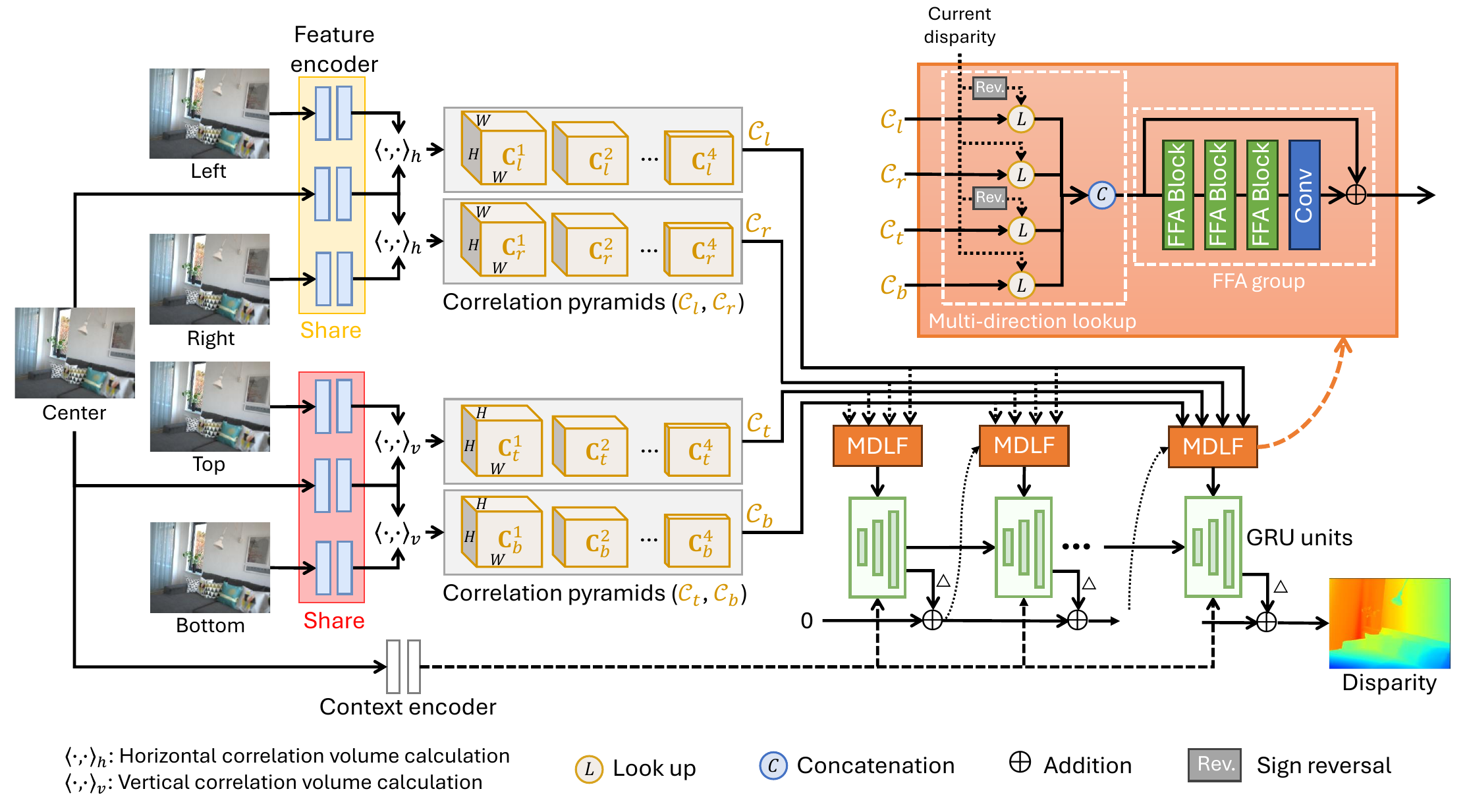}}
\end{center}
   \vspace{-10mm}
   \caption{The overall architecture of our QPDNet.}
\label{fig:network}
\end{figure*}

Figure~\ref{fig:network} presents the overall architecture of our QPDNet. We employ RAFT-Stereo~\cite{lipson2021raft} as our baseline architecture. We use the center image as the reference image and apply two feature encoders to extract the features $\{{\bf F}_l, {\bf F}_{c_{h}}, {\bf F}_r\}$ for horizontal matching between center-left and center-right image pairs and the features $\{{\bf F}_t, {\bf F}_{c_{v}}, {\bf F}_b\}$ for vertical matching between center-top and center-bottom image pairs, respectively. The dimension of each feature is ${H\times W \times N_F}$, where $H$ and $W$ represent the image height and width, respectively, and $N_F$ is the number of channels (specifically, $N_F=256$).
Subsequently, we calculate four correlation volumes $\{{\bf C}_l, {\bf C}_r, {\bf C}_t, {\bf C}_b\}$ as 
\begin{equation}
  {\bf C}_i=
  \begin{cases}
    <{\bf F}_{c_{h}}, {\bf F}_i>_h & \text{if $i\in \{l, r\}$,} \\
    <{\bf F}_{c_{v}}, {\bf F}_i>_v & \text{if $i\in \{t, b\}$,}
  \end{cases}
\end{equation}
where $<{\bf F}_{c_{h}}, {\bf F}_i>_h$ represents the same 3D correlation volume construction operation as RAFT-Stereo, where the inner product of the features from ${\bf F}_{c_{h}}$ and ${\bf F}_i$ is taken by horizontally shifting the feature map ${\bf F}_i$. Similarly, $<{\bf F}_{c_{v}}, {\bf F}_i>_v$ represents the 3D correlation volume construction operation by vertically shifting the feature map ${\bf F}_i$. Consequently, the dimension of the correlation volumes $\{{\bf C}_l, {\bf C}_r\}$ result in ${H\times W \times W}$, while the dimension of the correlation volumes $\{{\bf C}_t, {\bf C}_b\}$ result in ${H\times W \times H}$. Subsequently, the average pooling of the last dimension is applied three times for each correlation volume to construct a correlation pyramid with four scales, which is described as $\mathcal{C}_i = \{{\bf C}_i^1, {\bf C}_i^2, {\bf C}_i^3, {\bf C}_i^4\}$.

The derived four correlation pyramids $\{\mathcal{C}_l, \mathcal{C}_r, \mathcal{C}_t, \mathcal{C}_b\}$ are fed into our proposed MDLF module, as shown in the upper-right part of Fig~\ref{fig:network}, to produce a fused local correlation feature from all computed correlations in different directions and scales, as detailed in the next subsection.
Finally, akin to RAFT-Stereo, we initialize the disparity to zero and iteratively update the disparity through the GRU module, where the fused local correlation feature, the current disparity, and the context feature extracted from the center image by a context encoder are concatenated and fed into the GRU module.

We supervise our OPDNet in the same manner as RAFT-Stereo using the $L_1$ loss between the ground-truth and the predicted disparity maps in each GRU iteration as
\begin{equation}
\mathcal{L}=\sum_{j=1}^{N}\gamma^{N-j}\Vert{{\bf D}_{gt}-{\bf D}_{j}}\Vert_{1}   ,
\label{formula: loss}
\end{equation}
where $N$ represents the number of GRU iterations, ${\bf D}_{gt}$ and ${\bf D}_{j}$ denote the ground-truth disparity map and the predicted disparity map in iteration $j$, respectively, and $\gamma^{N-j}$ denotes the weight for each iteration, which increases with the number of iterations.

\subsection{Multi-Direction Lookup Fusion Module}

The upper-right part of Fig.~\ref{fig:network} shows our proposed MDLF module comprising two components: multi-direction lookup and feature fusion attention~(FFA) group.

The inputs to the MDLF module are the set of four correlation pyramids $\{\mathcal{C}_l, \mathcal{C}_r, \mathcal{C}_t, \mathcal{C}_b\}$. 
For each pyramid, local correlations are looked up based on the current disparity. Specifically, for the correlation volume of each scale in the pyramid, the length of $(2r+1)$ correlation value vector is extracted around the pixel position associated with the current disparity. Thus, by combining the correlation value vectors of all four scales, the length of $4\cdot(2r+1)$ vector is derived for each pixel. We further concatenate the looked-up $4\cdot(2r+1)$-length vectors from all four pyramids, resulting in a $16\cdot(2r+1)$-length vector for each pixel. By combining all the pixels, the local correlation feature map ${\bf F}_{cor}$ with the dimension of ${H\times W \times 16\cdot(2r+1)}$ is derived and fed into the next FFA group. In the case of the QP sensor, the amount of disparity is the same, but the direction is opposite for the center-left and the center-right disparities as well as the center-top and the center-bottom disparities. Thus, we reverse the sign of the current disparity when looking up the correlations in $\{\mathcal{C}_l, \mathcal{C}_t\}$ according to our definition of the disparity.

Subsequently, the derived local correlation feature map ${\bf F}_{Cor}$ is fed into the next FFA group, where channel-wise and pixel-wise attentions are employed to fuse the local correlations adaptively. Following~\cite{qin2020ffa}, we utilize an FFA group, comprising three FFA blocks and a final convolutional layer, as shown in Fig~\ref{fig:network}. 
Each FFA block with skip connection is composed as 
\begin{equation}
\begin{aligned}
\tilde{{\bf F}}_{cor} = ({\bf F}_{cor}+Conv({\bf F}_{cor})) \odot {\bf A}_c \odot {\bf A}_p+{\bf F}_{cor},
\end{aligned}
\label{formula: volume}
\end{equation}
where $\tilde{{\bf F}}_{cor}$ is the output feature of the FFA block, ${\bf A}_c$ is the channel-wise attention, and ${\bf A}_p$ is the pixel-wise attention. All of them have the same dimension as the input feature ${\bf F}_{cor}$ and the operation $\odot$ represents element-wise multiplication.  Specifically, ${\bf A}_c$ is formed by copying $1 \times 1 \times 16\cdot(2r+1)$-sized channel-wise attention to all the pixels. Similarly, ${\bf A}_p$ is formed by copying $H\times W \times 1$-sized pixel-wise attention to all the channels. Those attentions are element-wise multiplied with the feature map to adaptively fuse the local correlations in ${\bf F}_{cor}$ by learned attentions. To derive ${\bf A}_c$ and ${\bf A}_p$, we adopt the same manner as~\cite{qin2020ffa}.
The output from the final (i.e. the third) FFG block is then passed through a convolutional layer with skip connection as
\begin{equation}
\begin{aligned}
{\bf F}’_{cor} = Conv(\tilde{{\bf F}}_{cor})+{\bf F}_{cor},
\end{aligned}
\label{formula: volume}
\end{equation}
where ${\bf F}’_{cor}$ is the final output of the MDLF module.
\section{Experimental Results}
\subsection{Settings}
\label{Sec:Settings}
\textbf{Camera and Training Data Settings}. An Olympus OM-1 camera with a ZUIKO DIGITAL 25mm F1.8 lens was employed to capture real-world QP data. The camera settings include the F-number of 1.8 and the focal length of 25mm with the focus distance set at 4m. Due to the huge size of the full-resolution images (5184$\times$3888 pixels), the captured images were resized to one-third of their original size to facilitate the processing by the network. The pixel size was computed as the quotient of the sensor width and the horizontal resolution, where the sensor width is 0.0174m and the horizontal resolution after the post-resizing is~1728, resulting in the pixel size of $1.01e^{-5}$m.

For the generation of the synthetic QP dataset, we utilized all-in-focus images and depth maps provided in the Hypersim dataset~\cite{roberts2021}. During the generation process, the camera parameters identical to those of the real Olympus camera were applied. The depth range was set to 0.5-50m and only the scenes within this range were used. The minimal unit of the PSF radius was set to 0.01 pixels to precisely simulate the PSF kernels in high resolution. The generated synthetic QP dataset consists of 377 scenes of 1024$\times$768 pixels. The dataset was divided into 301 scenes for training, 38 scenes for validation, and the rest 38 scenes for testing, respectively. Additionally, a test set with noise was generated by adding Gaussian noise with a variance of 0.01 to evaluate the robustness against noise, which is inevitable in real data.

\noindent \textbf{Training Details}.
For training our proposed QPDNet, we set the gamma value in Eq.~\ref{formula: loss} as 0.9. The input images were randomly cropped to dimensions of $452\times452$. Regarding training parameters, the learning rate was set to $2e^{-4}$, the weight decay was configured at~$1e^{-5}$, and the batch size was set as 4. We iterated the GRU module eight times during both the training and the testing phases. The network was trained over a total of 265 epochs. The training took roughly 60 hours using a single NVIDIA GeForce RTX 4090 GPU.

\noindent \textbf{Compared Methods}. We compared our QPDNet with existing disparity estimation methods. Given the absence of methods specifically tailored for QP data, our comparison encompasses stereo methods and DP methods.

For the stereo methods, we selected RAFT-Stereo~\cite{lipson2021raft}, which is the baseline architecture of our QPDNet, and IGEV-Stereo~\cite{xu2023iterative}, which is another state-of-the-art method. To fairly compare the results with ours on the center-image-aligned disparity maps, we mixed the center-left and the center-right image pairs as the input data to train a single stereo network. Then, during the testing phase, the final center-image-aligned disparity map was derived by simply averaging the two disparity maps predicted from the center-left and the center-right image pairs.

For the DP methods, we compared MDD~\cite{punnappurath2020modeling}, which is an optimization-based method and estimates the blur kernel size map of the DP image, from which the disparity is derived with scale ambiguity. We also compared with SFBD~\cite{Kim2023Dualpixel}, which is a state-of-the-art learning-based method trained in a self-supervised manner. Because the training code of SFBD is not publicly available, we utilized the pre-trained SFBD model for disparity estimation. Furthermore, we performed supervised training of the SFBD model only using DP (i.e. center-left and center-right) images among our training data. During this supervised training process, we replaced the photometric loss of the original SFBD with the supervised $L_1$ loss, while maintaining the other kernel-split symmetry loss and smoothness loss as detailed in the original paper.

\noindent \textbf{Evaluation Metrics}. For the methods except for MDD that can directly estimate the disparity, we used the mean absolute error~(MAE), the root mean squared error~(RMSE), and $d_{0.5}$, $d_{1}$, and $d_{2}$ metrics, which describe the error percentage bigger than the thresholds of 0.5, 1.0, and 2.0 pixels, respectively. To compare with the MDD method which only can estimate the disparity with scale ambiguity, we adopted the affine invariant metrics of $AI(1)$ and $AI(2)$ in the MDD paper~\cite{punnappurath2020modeling}. The $AI(q)$ is defined as 
\begin{equation}
\begin{aligned}
AI(q) = \mathop{\min}\limits_{a,b} \left(\frac{\sum_{(x,y)}^{} \lvert {D}_{gt}(x,y)-(a \hat{D}(x,y)+b) \rvert^{q}}{N_I} \right)^{\frac{1}{q}},
\end{aligned}
\label{formula: volume}
\end{equation}
where $N_I$ is the total number of image pixels, $D_{gt}$ is the ground-truth disparity, and $\hat{D}$ is the estimated disparity. When $q=2$, $a$ and $b$ are obtained by a least squares method. When $q=1$, $a$ and $b$ are computed using iteratively re-weighted least squares.

\begin{table}
\vspace{3mm}
\begin{center}

 \setlength{\tabcolsep}{1.0mm}{
 \scalebox{0.9}{
  \begin{tabular}{ccc|ccccc|ccccc} 
  \hline

\multirow{2}{*}{Input} & \multirow{2}{*}{MDL} & \multirow{2}{*}{FFA} & \multicolumn{5}{c|}{w/o noise} & \multicolumn{5}{c}{w/ noise} \\ \cline{4-13}
&                      &                        & MAE  & RMSE  & d0.5 & d1 & d2 & MAE  & RMSE & d0.5 & d1 & d2 \\ \cline{1-13}

  CL,CR &  &  &0.030 & 0.163 & 0.840 & 0.368 & 0.137 & 0.148 & 0.510 & 4.932 & 2.613& 1.340 \\
  CLR & \checkmark &  &0.028 & 0.151 & 0.764 & 0.343 & 0.128 & 0.094 & 0.334 & 2.782 & 1.260 & 0.579 \\
  CLR & \checkmark & \checkmark &0.027 & 0.144 & 0.762 & 0.341 & 0.117 & 0.102 & 0.291 & 3.866 & 1.493 & 0.450 \\
  CLRTB & \checkmark & & 0.026 & 0.143 & 0.717 & 0.319 & 0.118 & 0.088 & 0.309 & 2.649 & 1.115 & 0.492 \\
  CLRTB & \checkmark & \checkmark & \textbf{0.025} & \textbf{0.142} & \textbf{0.703} & \textbf{0.317} & \textbf{0.116} & \textbf{0.074} & \textbf{0.264} & \textbf{2.129} & \textbf{0.956} & \textbf{0.366}\\
  \hline
  \end{tabular}}}
\end{center}
\vspace{-2mm}
\caption{The results of an ablation study.}
\label{table:ablation}
\end{table}

\subsection{Ablation Study}

To validate the efficacy of each component of our QPDNet, we conducted an ablation study. Table~\ref{table:ablation} summarizes the performance of various configurations. 
The first row represents the baseline method, which is a simple adoption of RAFT-Stereo~\cite{lipson2021raft} to DP data, as explained in Sec.~\ref{Sec:Settings}.  
The second row represents the method that adopts our multi-direction lookup~(MDL) to the center-left and the center-right image pairs (i.e., the DP sensor configuration). The third row represents the method that further adopts the FFA group after the MDL. This method is the DP sensor version of our proposed architecture. Comparing these three methods, we can confirm that the disparity estimation accuracy is improved from the first row to the third row, validating the effectiveness of MDL and FFA even for the DP sensor configuration.

The fourth row represents the method using the QP data as the inputs. This method applies our MDL to all of the center-left, the center-right, the center-top, and the center-bottom image pairs. The last row represents our final proposal (i.e., QPDNet) that applies both MDL and FFA to the QP data. From the results, we can confirm that our QPDNet provides the best performance, especially showing the robustness to noise. 

\subsection{Comparison with Previous Methods}

\begin{figure*}[t!]
\begin{center}
\centerline{\includegraphics[width=13cm]{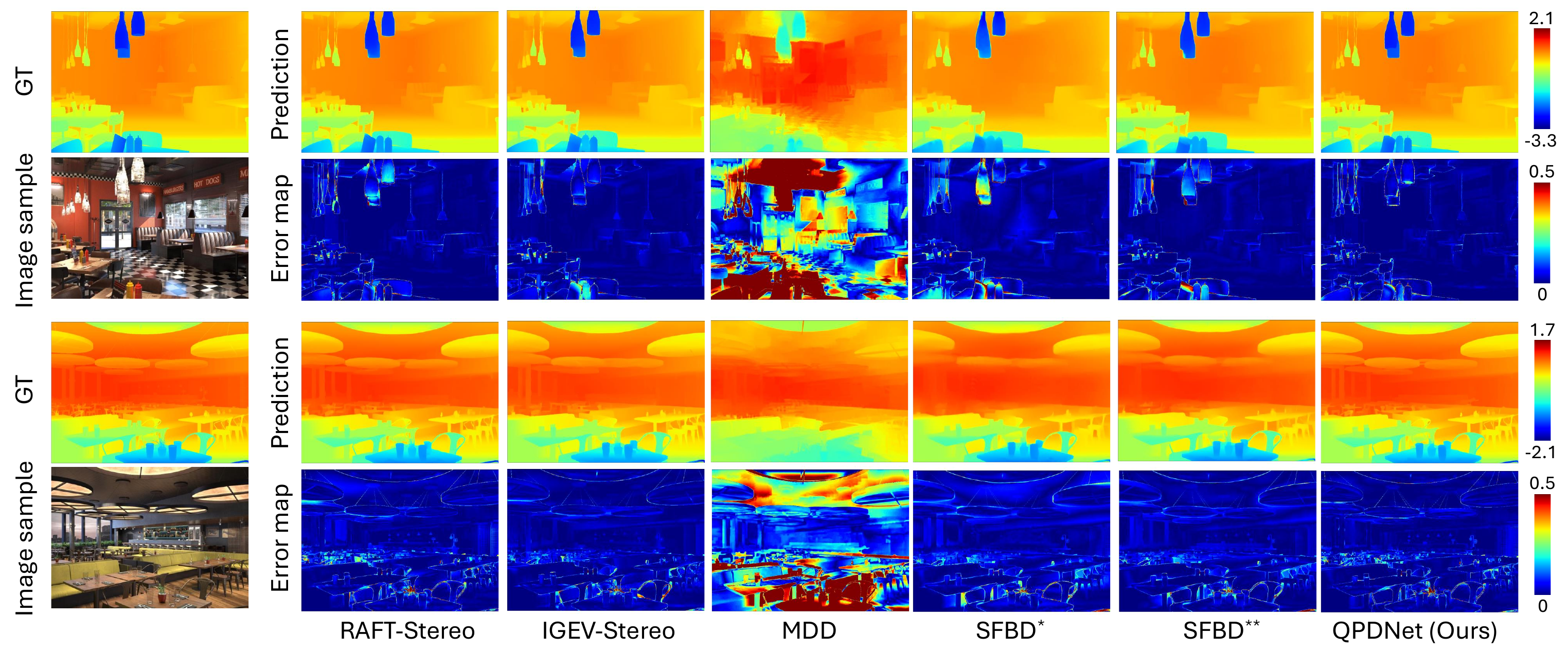}}
\end{center}
   \vspace{-10mm}
   \caption{Visual comparisons on the noise-free synthetic dataset.}
   \vspace{-2mm}
\label{fig:syn_result}
\end{figure*}

\textbf{Evaluation on Synthetic Data}. Table~\ref{table:others} and Fig.~\ref{fig:syn_result} show quantitative and visual results on the noise-free synthetic test dataset. In the noise-free testing results, the pre-trained SFBD$^*$ shows worse $AI(q)$ performance than the optimization-based method of MDD. In contrast, our trained SFBD$^{**}$ exhibits improved performance compared to the pre-trained SFBD$^*$. RAFT-Stereo and IGEV-Stereo, which were also directly trained on our training set, show better performance than SFBD$^{**}$. Our proposed QPDNet outperforms the other methods on all evaluated metrics for the noise-free case.

\begin{figure*}[t!]
\begin{center}
\centerline{\includegraphics[width=13cm]{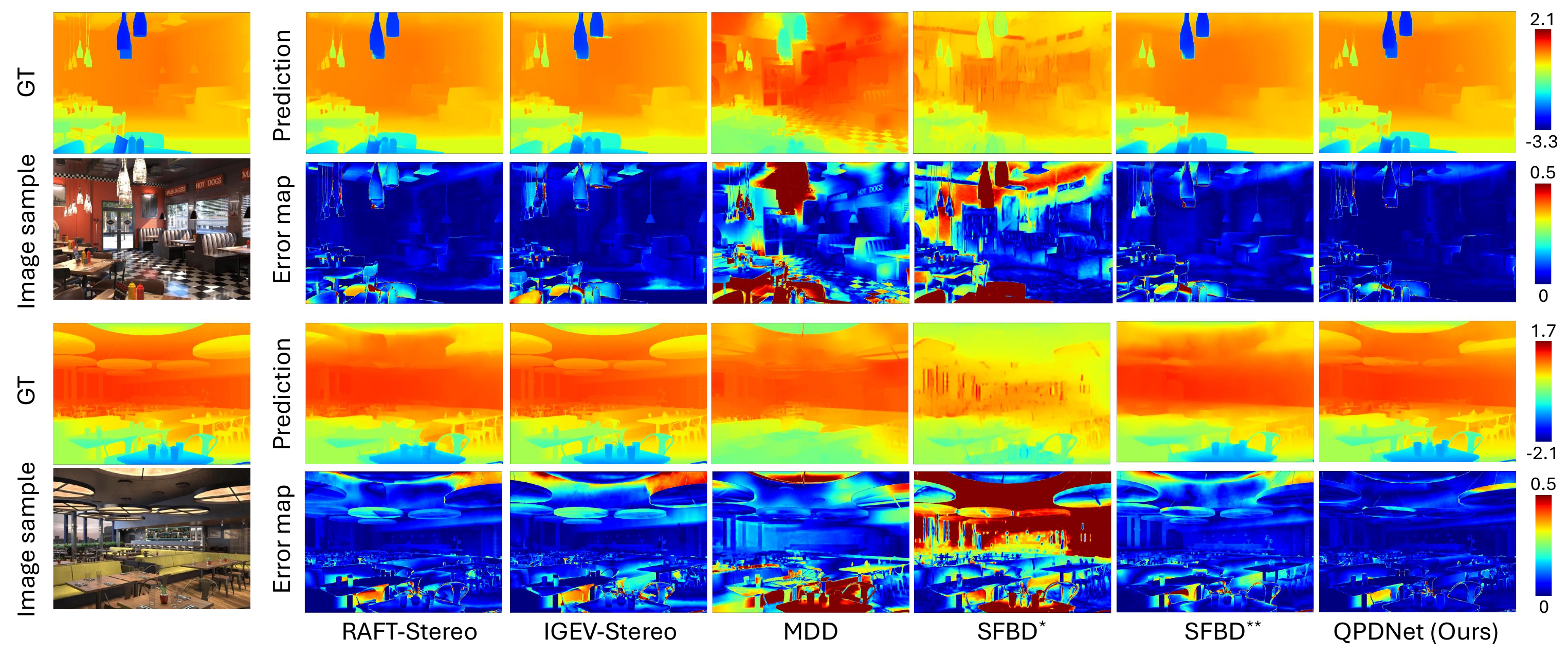}}
\end{center}
   \vspace{-10mm}
   \caption{Visual comparisons on the synthetic data with Gaussian noise.}
\label{fig:syn_result_noise}
\end{figure*}

As the disparities are quite small in the QP data, the differences among various methods are subtle under the noise-free condition. However, when subjected to testing data with noise as shown in Table~\ref{table:others_with_noise}, the robustness of each method becomes more evident. As illustrated in Table~\ref{table:others_with_noise}, the observed tendencies remain similar to those of the noise-free results, though SFBD$^{**}$ outperforms RAFT-Stereo and IGEV-Stereo for the noisy case. Notably, our QPDNet maintains its superiority, demonstrating enhanced performance and robustness. Furthermore, visual comparisons depicted in Fig.~\ref{fig:syn_result_noise} highlight the superior performance of our QPDNet to the other methods.

\begin{table}[t!]
\vspace{3mm}
\begin{center}

\setlength{\tabcolsep}{2.0mm}{
\begin{tabular}{c|ccccccc}
\hline
\multicolumn{1}{c|}{Methods} & MAE & RMSE & $d_{0.5}$ & $d_1$  & \multicolumn{1}{l|}{$d_2$} & AI(1) & AI(2) \\ \hline
RAFT-Stereo~\cite{lipson2021raft} & 0.030 & 0.163 & 0.840 & 0.368 & \multicolumn{1}{l|}{0.137} & 0.030 & 0.082 \\
IGEV-Stereo~\cite{xu2023iterative}  & 0.035 & 0.179 & 0.854 & 0.376 & \multicolumn{1}{l|}{0.165} & 0.035 & 0.093 \\ 
MDD~\cite{punnappurath2020modeling}  & -- & -- & -- & -- & \multicolumn{1}{l|}{--} & 0.208 & 0.324 \\
SFBD$^*$~\cite{Kim2023Dualpixel}  & 0.290 & 0.754 & 7.315 & 3.470 & \multicolumn{1}{l|}{0.876} & 0.262 & 0.430 \\
SFBD$^{**}$~\cite{Kim2023Dualpixel}  & 0.038 & 0.187 & 1.085 & 0.472 & \multicolumn{1}{l|}{0.177} & 0.041 & 0.107 \\
QPDNet~(Ours) & \textbf{0.025} & \textbf{0.142} & \textbf{0.703} & \textbf{0.317} & \multicolumn{1}{l|}{\textbf{0.116}} & \textbf{0.025} & \textbf{0.074} \\ 
\hline
\end{tabular}}
\end{center}
\caption{The quantitative comparisons on the noise-free synthetic dataset. SFBD$^*$ indicates the utilization of pre-trained weights, while SFBD$^{**}$ refers to the model that was trained on our training data in a supervised manner.}
\label{table:others}
\end{table}

\begin{table}[t!]
\vspace{3mm}
\begin{center}
\setlength{\tabcolsep}{2.0mm}{
\begin{tabular}{c|ccccccc}
\hline
\multicolumn{1}{c|}{Methods} & MAE & RMSE & $d_{0.5}$ & $d_1$  & \multicolumn{1}{l|}{$d_2$} & AI(1) & AI(2) \\ \hline
RAFT-Stereo~\cite{lipson2021raft} & 0.148 & 0.510 & 4.932 & 2.613 & \multicolumn{1}{l|}{1.340} & 0.144 & 0.271 \\
IGEV-Stereo~\cite{xu2023iterative}  & 0.141 & 0.452 & 4.648 & 2.056 & \multicolumn{1}{l|}{\textbf{0.355}} & 0.136 & 0.256 \\ 
MDD~\cite{punnappurath2020modeling}  & -- & -- & -- & -- & \multicolumn{1}{l|}{--} & 0.254 & 0.367 \\
SFBD$^*$~\cite{Kim2023Dualpixel}  & 0.419 & 0.837 & 23.266 & 8.764 & \multicolumn{1}{l|}{3.933} & 0.357 & 0.528 \\
SFBD$^{**}$~\cite{Kim2023Dualpixel}  & 0.110 & 0.276 & 3.011 & 1.046 & \multicolumn{1}{l|}{0.309} & 0.103 & 0.191 \\
QPDNet~(Ours) & \textbf{0.074} & \textbf{0.264} & \textbf{2.129} & \textbf{0.956} & \multicolumn{1}{l|}{0.366} & \textbf{0.072} & \textbf{0.153} \\ 
\hline
\end{tabular}}
\end{center}
\caption{The quantitative comparisons on the synthetic dataset with Gaussian noise. SFBD$^*$ indicates the utilization of pre-trained weights, while SFBD$^{**}$ refers to the model that was trained on our training data in a supervised manner.}
\vspace{-1mm}
\label{table:others_with_noise}
\end{table}

\begin{figure*}[t!]
\begin{center}
\centerline{\includegraphics[width=12.5cm]{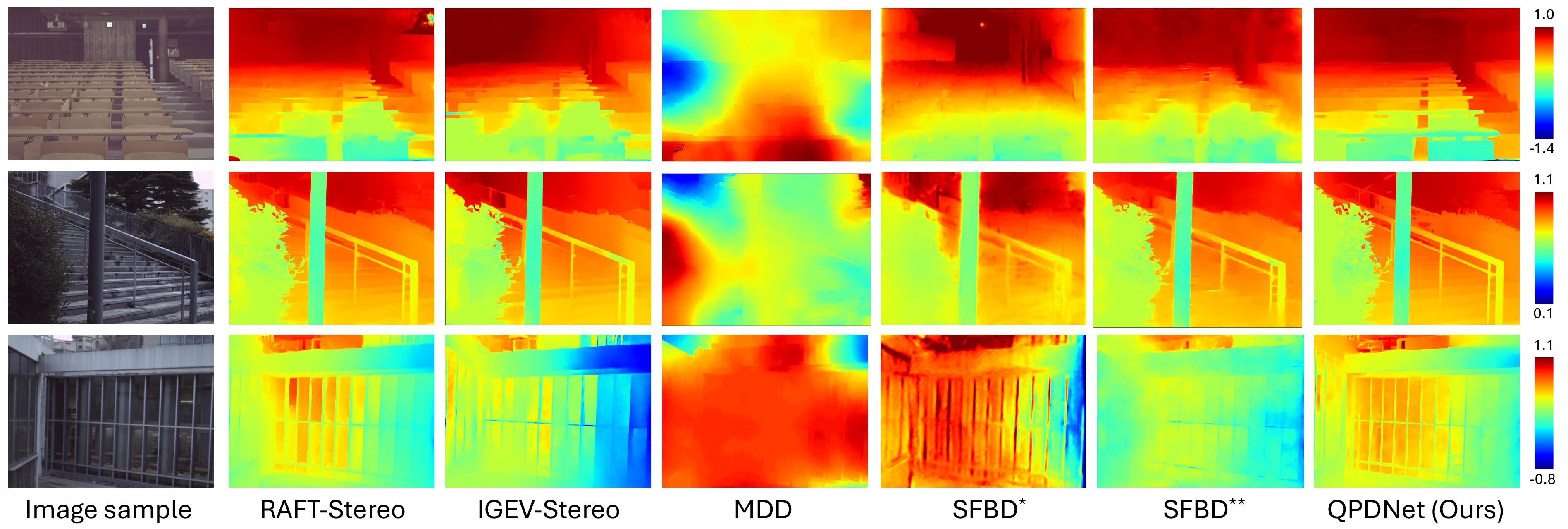}}
\end{center}
   \vspace{-10mm}
   \caption{Visual comparisons on real-scene data.}
\label{fig:real_result}
\end{figure*}

\noindent \textbf{Evaluation on Real Scene Data}. To further evaluate the robustness of our QPDNet, we conducted evaluations using real-world data. Because obtaining ground-truth disparity maps aligned with the real QP sensor is difficult, our evaluation relies on qualitative visual comparisons. Figure~\ref{fig:real_result} illustrates the visual outcomes derived from the real-world data. Overall, the optimization-based MDD method struggles to produce accurate disparity maps. In the second row of the figure, all the methods except MDD produce satisfactory estimation results. However, in the first and third rows, it is evident that our QPDNet consistently predicts better and sharper disparity maps in real-world scenarios.

\section{Conclusion}

In this paper, we have proposed a novel network, QPDNet, to estimate the disparity using a QP sensor. Our proposed QPDNet leverages the multi-directional disparity information provided by the QP data. Specifically, we have constructed multi-directional stereo-matching correlation volumes and effectively fused them with a novel MDFL module. To train and test OPDNet, we have constructed a synthetic QP dataset. Through extensive experiments, we have validated the effectiveness of our OPDNet. One of the possible future works includes the extension of QPDNet to a self-supervised method to better generalize to real-scene data, by using the data obtained by a real QP camera including non-ideal properties, such as noise, distorted PSFs, and unbalanced brightness.

\bibliography{egbib}
\end{document}